\documentclass[letterpaper, 10 pt, conference]{ieeeconf}  

\IEEEoverridecommandlockouts                              

\usepackage{amsmath} 

\usepackage{graphics} 
\usepackage{epsfig} 
\usepackage{times} 
\usepackage{amsmath} 
\usepackage{amssymb}  
\usepackage{hyperref} 
\usepackage{xcolor} 
\usepackage{arydshln} 
\usepackage{color} 
\usepackage{booktabs}
\usepackage{hyperref}
\usepackage{float}
\usepackage{multirow}
\usepackage{pifont}
\usepackage{bm}
\usepackage{tikz}
\usepackage{tabularx}
\usepackage{tikz-3dplot}
\usepackage{multirow}
\usepackage{mathtools}
\usepackage{graphicx}
\usepackage{amsfonts}
\usepackage{units}
\newcommand{\cmark}{\ding{51}}%
\newcommand{\xmark}{\ding{55}}%
\usepackage{balance}
\usepackage{cite}

\newcommand{\rotateRPY}[3]
{   \pgfmathsetmacro{\rollangle}{#1}
    \pgfmathsetmacro{\pitchangle}{#2}
    \pgfmathsetmacro{\yawangle}{#3}

    \pgfmathsetmacro{\newxx}{cos(\yawangle)*cos(\pitchangle)}
    \pgfmathsetmacro{\newxy}{sin(\yawangle)*cos(\pitchangle)}
    \pgfmathsetmacro{\newxz}{-sin(\pitchangle)}
    \path (\newxx,\newxy,\newxz);
    \pgfgetlastxy{\nxx}{\nxy};

    \pgfmathsetmacro{\newyx}{cos(\yawangle)*sin(\pitchangle)*sin(\rollangle)-sin(\yawangle)*cos(\rollangle)}
    \pgfmathsetmacro{\newyy}{sin(\yawangle)*sin(\pitchangle)*sin(\rollangle)+ cos(\yawangle)*cos(\rollangle)}
    \pgfmathsetmacro{\newyz}{cos(\pitchangle)*sin(\rollangle)}
    \path (\newyx,\newyy,\newyz);
    \pgfgetlastxy{\nyx}{\nyy};

    \pgfmathsetmacro{\newzx}{cos(\yawangle)*sin(\pitchangle)*cos(\rollangle)+ sin(\yawangle)*sin(\rollangle)}
    \pgfmathsetmacro{\newzy}{sin(\yawangle)*sin(\pitchangle)*cos(\rollangle)-cos(\yawangle)*sin(\rollangle)}
    \pgfmathsetmacro{\newzz}{cos(\pitchangle)*cos(\rollangle)}
    \path (\newzx,\newzy,\newzz);
    \pgfgetlastxy{\nzx}{\nzy};
}

\usepackage{tabularx}
\usepackage{tikz}
\usepackage{tikz-3dplot}
\usepackage{pgfplots}
\usepackage{pgfplotstable}
\usetikzlibrary{bending, external, plotmarks, shapes, pgfplots.statistics, pgfplots.colorbrewer, positioning}
\pgfplotsset{compat=1.16} 

\newcolumntype{C}{>{\centering\arraybackslash}X}
\newcolumntype{x}[1]{>{\centering\let\newline\\\arraybackslash\hspace{0pt}}p{#1}}
\definecolor{matlab1}{rgb}{0.00000,0.44700,0.74100}
\definecolor{matlab2}{rgb}{0.85000,0.32500,0.09800}
\definecolor{matlab3}{rgb}{0.92900,0.69400,0.12500}
\definecolor{matlab4}{rgb}{0.49400,0.18400,0.55600}
\definecolor{matlab5}{rgb}{0.4660, 0.6740, 0.1880}
\definecolor{matlab6}{rgb}{0.3010, 0.7450, 0.9330}
\definecolor{matlab7}{rgb}{0.6350, 0.0780, 0.1840}
\definecolor{matlab8}{rgb}{0.8, 0.8, 0}
\definecolor{matlab9}{rgb}{0.5, 0.5, 0.5}
\definecolor{verylightgray}{rgb}{0.98,0.98,0.98}

\pgfdeclarelayer{bg}   

\definecolor{somegray}{rgb}{0.5, 0.5, 0.5}
\newcommand{\darkgrayed}[1]{\textcolor{somegray}{#1}}
\makeatletter
\newcommand*\titleheader[1]{\gdef\@titleheader{#1}}
\AtBeginDocument{%
  \let\st@red@title\@title
  \def\@title{%
    \vskip-3em
    \bgroup\normalfont\large\centering\@titleheader\par\egroup
    \vskip1.5em\st@red@title}
}
\makeatother
\titleheader{\darkgrayed{This paper has been accepted for publication at the\\
IEEE International Conference on Robotics and Automation (ICRA), Yokohama 2024
\copyright IEEE}}
\title{\LARGE \bf
Contrastive Learning for Enhancing Robust Scene Transfer in Vision-based Agile Flight
}

\author{Jiaxu Xing$^{*}$, Leonard Bauersfeld$^{*}$, Yunlong Song, Chunwei Xing, and Davide Scaramuzza
    \thanks{$^{*}$ These authors contributed equally.
    The authors are with the Robotics and Perception Group, Department of Informatics, University of Zurich, and Department of Neuroinformatics, University of Zurich and ETH Zurich, Switzerland (\protect\url{http://rpg.ifi.uzh.ch}). 
    This work was supported by the European Union’s Horizon 2020 Research and Innovation Programme under grant agreement No. 871479 (AERIAL-CORE), and the European Research Council (ERC) under grant agreement No. 864042 (AGILEFLIGHT).
    }
}
\begin{document}
\makeatletter
\maketitle
\thispagestyle{empty}
\pagestyle{empty}
\begin{abstract}
Scene transfer for vision-based mobile robotics applications is a highly relevant and challenging problem. The utility of a robot greatly depends on its ability to perform a task in the real world, outside of a well-controlled lab environment. 
Existing scene transfer end-to-end policy learning approaches often suffer from poor sample efficiency or limited generalization capabilities, making them unsuitable for mobile robotics applications. 
This work proposes an adaptive multi-pair contrastive learning strategy for visual representation learning that enables zero-shot scene transfer and real-world deployment. 
Control policies relying on the embedding are able to operate in unseen environments without the need for finetuning in the deployment environment. 
We demonstrate the performance of our approach on the task of agile, vision-based quadrotor flight. Extensive simulation and real-world experiments demonstrate that our approach successfully generalizes beyond the training domain and outperforms all baselines. \textbf{Video: }\href{https://youtu.be/4A4YyPgEWD8}{https://youtu.be/4A4YyPgEWD8}
\end{abstract}
\section{Introduction}
The ideal mobile robot could perform any task, anywhere, and without relying on external sensors\textemdash just like us humans. However, even the most advanced autonomous mobile robots only surpass human performance in well-controlled conditions on a narrowly defined task~\cite{kaufmann2023nature}.
Many recent works on multi-task learning~\cite{fifty_efficiently_2021, shridhar2022perceiver, rahmatizadeh2018vision} address one part of this problem, namely how a robot can learn to perform different tasks in one environment. 
However, a mobile robot's utility also strongly depends on its ability to perform the same task outside of a known environment in the real world. 
To possess such \emph{scene transfer} capabilities, the robot requires a robust perception and control system that operates consistently across different environments. 

\begin{figure}[t!]
    \centering
    \includegraphics[width=0.99\linewidth]{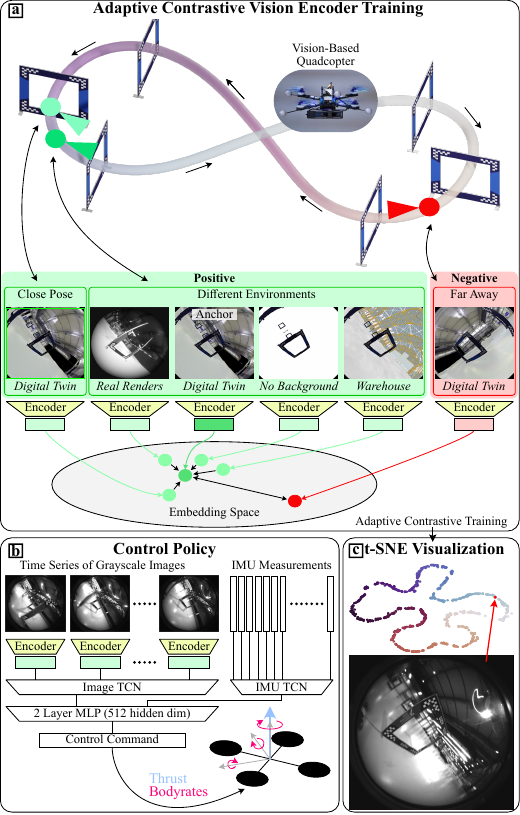}
    \vspace*{-18pt}
    \caption{We train a vision encoder using our proposed adaptive contrastive learning strategy. Positive examples are sampled from different environments and nearby points and negative examples are from far-away track segments, as shown in (a). 
    Then the action network controlling the robot has access to a history of vision embeddings as well as IMU measurements. The action net predicts the control commands for the mobile robot, e.g. thrust and body rate commands for a quadrotor, as shown in (b).
    Our adaptive contrastive learning embeds the images into a self-consistent and scene-invariant feature space, shown via a t-SNE~\cite{van2008visualizing} visualization in (c).}
    \label{fig:method}
    \vspace*{-23pt}
\end{figure}

In terms of the well-established paradigm \cite{zhou2019cv, sax2020visprior, driess_reinforcement_2022} to compress the perception of the environment into an intermediate representation and then use this for the action (control) network, the challenge can be rephrased as follows: how can we learn a task-related perception embedding that is invariant to the environment in which the task needs to be performed?
Such a perception embedding removes the need for finetuning the action network in a specific environment and thus is an important step towards the zero-shot deployment of a mobile robot in a new environment and the real world. 

For agile aerial vehicles, it is especially difficult to learn a vision representation that enables robust scene transfer. 
These robots can move freely in 3D space, and thus, even in the same environment, the observations can be vastly different. 
Additionally, there are few constraints on how the environment will look as both nearly textureless or feature-rich indoor and outdoor environments are possible.
The action network must control a system that moves at high speed and requires constant action to not crash into obstacles or the environment.
Finetuning the policy in the real world is difficult due to safety concerns and because finetuning requires the policy to perform well enough to collect data and not crash the robot immediately.

These challenges present a stark contrast to prior work on scene transfer. One family of works addresses scene transferability for vision-based object manipulation and shows impressive robustness to the clutter in the scene~\cite{zeng2022robotic, gato} and background invariance~\cite{schwarz2018rgb}.
In the context of mobile robot navigation, end-to-end vision-based control has primarily been explored in settings where the training and testing environment are the same (e.g., simulation~\cite{Wurman22Nature, yunlong2021sony, yunlong2021super, muller2018teaching}) or as a way to bridge the sim2real gap. 
The latter can be done using a handcrafted intermediate representation that is task- and environment-specific~\cite{kaufmann2023nature, kaufmann2020deep, xiao2022motioncontrol}. 

However, the existing approaches mentioned do not effectively tackle the challenge of scene transfer. Achieving vision-based end-to-end control for mobile robot navigation that generalizes beyond the training environment remains an unsolved problem. The significant challenge lies in obtaining a suitable visual embedding for action learning, with proposed solutions encompassing both world models~\cite{gato} and task-specific models~\cite{schwarz2018rgb, laskin2020curl}.
One key aspect is ensuring that the learned visual embedding is both environment-agnostic and task-relevant. Addressing this concern, contrastive learning emerges as an ideal approach for aligning the acquired embeddings. Contrastive learning serves as a robust framework for generating representations that adhere to similarity constraints within a dataset, typically structured around pairs of similar and dissimilar samples. Notably, contrastive learning has showcased promising performance on various robot learning tasks~\cite{laskin2020curl, messikommer2024contrasitve}.

In this work, we present an adaptive contrastive learning approach (shown in Fig.~\ref{fig:method}) to train a vision encoder that outputs a task-related but environment-independent embedding and that generalizes to the real world without finetuning. 
This advance is made possible by using i) a multi-pair contrastive loss combined with a novel formulation that replaces the constant temperature~\cite{frosst2019analyzing} approach with an adaptive temperature that captures the similarity of the contrastive pairs. ii) a novel contrastive pair sampling strategy to enhance the embedding similarity among various scenes, hence forcing the network to learn task-related features.
The vision encoder is paired with a small action net, trained via privileged imitation learning~\cite{Chen19corl} purely in simulation.
We present extensive simulation and real-world experiments, comparing our approach with world-model-type approaches and task-specific vision encoders.
Our approach outperforms all baselines and transfers to unseen environments where the action net has never been trained.

\section{Related Work}

\subsection{End-to-end policy learning} 
End-to-end sensorimotor policy learning aims to directly predict actions from raw sensor measurements, such as RGB~\cite{levine2016end} or depth images~\cite{loquercio2021learning, agarwal2022legged}. 
Contrary to traditional methods, such as~\cite{xing2023autnomous}, end-to-end approaches often operate without the requirement of environmental understanding~\cite{sun2022see, oleynikova2017voxblox, Rosinol20icra}, accurate state estimation~\cite{hanlon2024active, forster2014svo}, or motion planning~\cite{campos2021orb, forster2014svo}. 
These approaches rely less on handcrafted features or system redundancy. 
Regarding system control, recent work demonstrates the advantages of end-to-end approaches over classical optimal control in autonomous racing~\cite{song2023reaching}.
However, end-to-end approaches commonly face challenges related to poor sample efficiency and limited ability for scene transfer~\cite{ren2021scene}, and existing approaches try to overcome this problem using handcrafted intermediate representations~\cite{kaufmann2020deep, xiao2022motioncontrol, radosavovic2022real} to enable robust scene transfer.

\subsection{Visual pre-training for robotics}
Recently, visual pre-training has shown great success for various robotics tasks using learned action-centric representations~\cite{radosavovic2022real, yen2020learning, zhan2022pretrain}. Pretrained large-scale models, such as CLIP~\cite{radford2021learning} and masked autoencoder~\cite{mae2022kh}, have demonstrated significant potential in terms of generalization across various tasks for robotics~\cite{radosavovic2022real, shridhar2021cliport}. These models are trained in a supervised fashion on large datasets and have promising capabilities in adapting to diverse scenarios and tasks~\cite{shridhar2021cliport}. At the same time, self-supervised contrastive pretraining~\cite{oord2018representation, messikommer2024contrasitve} and contrastive reinforcement learning (RL) have also shown notable advancements in enhancing the performance of policy learning~\cite{qiu2022ucb, liu2021return}. 
In the context of this work, especially two works on contrastive RL are important, as our contrastive loss is inspired by them. In~\cite{eysenbach2022contrastive, correia2022multi}, a contrastive RL strategy was used to ensure that the representations of future states are closer than the representations of random states. Specifically, the positive and negative pairs are chosen based on the state trajectory of the manipulator. In the following, we refer to this strategy as \emph{State CL}. In~\cite{laskin2020curl} an image-based contrastive RL approach is proposed where the positive pairs are generated as augmentations of the same image. This image-based contrastive RL approach demonstrated a significant enhancement in sample efficiency for action policy learning. We refer to this as \emph{Aug CL}.

It is noteworthy that while the aforementioned approaches have demonstrated impressive achievements in various robotics tasks, none of them have shown significant capabilities for scene transfer in the domain of mobile robot navigation.
Our proposed contrastive learning approach focuses on enhancing scene transfer capabilities by using an adaptive contrastive learning approach and novel contrastive pair formation strategy. The resulting scene-consistent representation makes it particularly well-suited for mobile robot navigation tasks.

\section{Methodology}
Our goal is to learn a sensorimotor policy capable of directly mapping raw onboard images and proprioception (e.g. IMU measurements) into quadcopter control commands (thrust and body rates~\cite{kaufmann2022benchmark}). The design choice to incorporate IMU measurements for agile robot control is driven by the high sensor frequency at around 200Hz~\cite{kaufmann2020deep}. To evaluate the performance of the proposed method, we chose the task of autonomous, vision-based drone racing~\cite{hanover2023autonomous}. The goal is to navigate through a sequence of predefined racing gates in minimum time. Fig.~\ref{fig:method}a exemplarily shows 
the \emph{figure~8} racetrack along with the onboard camera images from various simulated and real environments. Drone racing is a challenging control task due to the high accelerations, non-linear aerodynamic effects, and low-latency requirements.

We follow the paradigm of using a vision-encoder which maps the perception input into an embedding space (Fig.~\ref{fig:method}a). The control policy then uses a time-series of such embeddings together with proprioceptive measurements as an input to output the control commands (Fig.~\ref{fig:method}b). 
We propose a ResNet-based vision encoder trained via multi-pair adaptive contrastive learning. The encoder is trained on images of a drone flying around the race track in multiple environments and the images require a pose (position and orientation) annotation. Additionally, we use the task-specific track progress to disambiguate potential self-intersections of the tracks. Collecting the images required to train the vision encoder can be done in many ways, e.g. in simulation, manually or automatically, and the pose annotation can be obtained from the simulation, motion-capture systems or structure-from-motion (SfM) approaches~\cite{schoenberger2016mvs, schoenberger2016sfm}. After the vision encoder has been trained it is frozen. 

To train the action network we use privileged imitation learning~\cite{learningbycheating} approach in a DAgger setting~\cite{ross2011reduction}. First, a state-based teacher policy is trained via deep reinforcement learning~\cite{song2021autonomous}. 
We use the imitation learning approach~\cite{torabi2018behavioral, ross2011reduction}, to distill knowledge from the teacher policy into the student policy. The vision-based student policy exclusively relies on image embeddings from the vision encoder and raw IMU measurements. 



\subsection{Adaptive Contrastive Learning}
\label{subsec:contrastive}
First, we will discuss our proposed contrastive training pipeline in detail and then provide a brief overview of the state-based teacher training and vision-based student training.
Given a set of $n$ different scenes $\mathcal{S} = \left\{S_1, \dots, S_n\right\}$, the primary objective is to train a vision encoder $\phi_{\bm\theta}(\cdot)$ with learnable parameters $\bm{\theta}$ that demonstrate both \emph{intra-scene} and \emph{inter-scene} consistency. 

\subsubsection{Intra-scene consistency} Within a scene $S_i$, consider a reference (subsequently referred to as \emph{anchor}) image $I_{\textrm{anc}}$ at a certain track progress $p_{\textrm{anchor}}$ (i.e. relative completion of the lap). The objective is to learn a representation that demonstrates similarity for nearby (w.r.t. track progress) images and dissimilarity for far images (see Fig.~\ref{fig:method}a, \emph{embedding space}). This ensures a descriptive embedding for the subsequent action policy learning process. The objective can be formulated as a contrastive learning problem, where the soft-nearest neighbor (SNN) loss~\cite{frosst2019analyzing} $\mathcal{L}_{\textrm{snn}}$ is minimized for each datapoint within a batch:
\begin{equation}
    \mathcal{L}_{\textrm{snn}} \!=\! -\log\frac{\sum_{p\in \mathcal{P}} \exp\big[-\textrm{sim}(\phi_{\bm{\theta}}(I_{\textrm{anc}}), \phi_{\bm{\theta}}(I_{\textrm{p}})) / \tau\big]}
   {\sum_{i\in \mathcal{N}\cup\mathcal{P}} \exp\!\big[-\textrm{sim}(\phi_{\bm{\theta}}(I_{\textrm{anc}}), \phi_{\bm{\theta}}(I_{\textrm{i}})) / \tau\big]}\!\!\!
   \label{eq:snn}
\end{equation}
Here $\mathcal{P}$, $\mathcal{N}$ refer to the sets of positive/negative samples which are samples of close/far progress (w.r.t. the anchor sample) within the track. $I_p$, $I_n$ represent the images perceived at the corresponding progress. $\textrm{sim}(a,b)$ represents the cosine similarity between two embedding vectors $a$ and $b$. 
The additional parameter $\tau$, commonly referred to as \emph{temperature}, can be used to guide the learned distribution in the embedding space. In~\cite{wang2021understanding} this effect is analyzed in detail, and the authors find that the selection of temperature significantly affects the quality of the learned representations.

We take these findings as an inspiration and augment the contrastive learning loss to a novel, adaptive temperature formulation which replaces the constant $\tau$. Given the nature of most mobile robot navigation tasks, we leverage pose information to judge how similar positive and negative pairs are. 
We define a similarity measurement for the robot pose $\bm{x} = \left[\bm{p}_{\textrm{norm}}, \bm{q}\right]$, where $\bm{p}_{\textrm{norm}}$ is the [0, 1] normalized position, $\bm{q}$ is the rotation quaternion, and $\alpha(\cdot,\cdot)$ calculates the angle between two quaternions: \vspace*{-6pt}
\begin{equation}
    \tau(\bm{x}_1, \bm{x}_2) = \tau(\bm{p}_1, \bm{q}_1, \bm{p}_2, \bm{q}_2) = \left|\bm{p}_1 - \bm{p}_2\right|_1 \cdot \frac{\alpha(\bm{q}_1, \bm{q}_2)}{\pi}
    \label{eq:similarity}
\end{equation}
Intuitively, if two poses are extremely similar, the adaptive $\tau$ ensures any dissimilarity in their embeddings is heavily weighted in eq.~(\ref{eq:snn}). 





\subsubsection{Inter-scene consistency} Apart from self-consistency within a scene, we want to improve the consistency across different scenes, as this property is key for scene transfer. For the drone racing example at hand, the embedding for the same pose should not depend on the background where the drone flies.
To achieve this, we sample random scene changes and image augmentations during training (see Fig.~\ref{fig:method}a, different environments). The positive and negative pairs are randomly sampled from any scene in $\mathcal{S}$ and augmented through contrast and brightness changes.
The vision encoder should enforce the images perceived at similar progress from different environments to have similar embeddings.
More specifically, the network training is formulated as:
\begin{equation}
    \operatorname*{argmin}_{\bm{\theta}} \; \mathop{\mathbb{E}}_{X_{\textrm{scene}}, X_{\textrm{aug}}}\left[
\mathcal{L}_{\textrm{snn}}(\bm{\theta};~I(x_{\textrm{scene}}, x_{\textrm{aug}}))
    \right]
\end{equation}
$ X_{\textrm{scene}}$, $X_{\textrm{aug}}$ are two random variables controlling scenes and data augmentation randomization, and $x_{\textrm{scene}}, x_{\textrm{aug}}$ are the corresponding realizations.
In our setting, we fully randomized the image in multiple scenes uniformly for all the contrastive pairs and applied an augmentation with a fixed probability.

\subsection{Action Net Learning}

\subsubsection{Teacher policy learning}
The state-based teacher policy takes the state observations $\mathbf o_t$ as input and outputs control commands. At each timestep, the policy has access to an observation from the environment which contains the current robot state (position, orientation, velocity, body rates) and the relative position to the next waypoint to be passed. 
The teacher policy is trained using model-free reinforcement learning approach~\cite{schulman2017proximal}, and we adapt the reward formulation based on~\cite{song2021autonomous, yunlong2020flightmare}. 
The reward $r_t$ at time $t$ is formulated as 
\begin{equation}
    r_t = r_t^{\textrm{prog}} + r_t^{\textrm{perc}} - r_t^{\textrm{act}} -r_t^{\textrm{br}} + r_t^{\textrm{pass}} - r_t^{\textrm{crash}}
\end{equation}
where $r_t^{\textrm{prog}}$ rewards progress towards the next gate to be
passed~\cite{yunlong2021racing}, $r_t^{\textrm{perc}}$ encodes perception awareness by adjusting the vehicle’s attitude such that the optical axis of its camera points towards the next gate’s center, $r_t^{\textrm{act}}$ penalizes action changes from the last time step, $r_t^{\textrm{br}}$ penalizes body rates and consequently reduces motion blur, $r_t^{\textrm{pass}}$ is a binary reward that is active when the robot successfully passes the next gate, $r_t^{\textrm{crash}}$ is a binary penalty that is only active when a collision happens, which also ends the episode. 
The reward components are formulated as follows:
\begin{equation}
    \begin{aligned}
        r_t^{\textrm{prog}} & = \lambda_1(d_{\textrm{Gate}}(t-1) - d_{\textrm{Gate}}(t)) \\
        r_t^{\textrm{perc}} & = \lambda_2 \exp(\lambda_3 \cdot \delta_{cam}^4) \\
        r_t^{\textrm{act}} & = \lambda_4 \lVert\bm{a}_t - \bm{a}_{t-1}\rVert \\
        r_t^{\textrm{br}} & = \lambda_6 \lVert\bm{\omega}_{\mathcal{B},t}\rVert \\
        r_t^{\textrm{pass}} & =  10.0 \quad\text{if robot passes the next gate} \\
        r_t^{\textrm{crash}} & = -5.0 \quad\text{if robot crashes (gates, ground) } \\
    \end{aligned}
\end{equation}
Here $d_{\textrm{Gate}}(t)$ denotes the distance from the robot's center of mass to the center of the next gate to pass, $\delta_{\textrm{cam}}$ is the angle between the camera's optical axis and the direction towards the center of the next gate. $\bm{a}$ represents the control command, and $\bm{\omega}$ the body rate. $\lambda_1, \lambda_2, \lambda_3, \lambda_4, \lambda_5$ are different hyperparameters.

\subsubsection{Student policy learning}
As shown in Fig.~\ref{fig:method}b the vision-based student policy takes a sequence (history length is 12 timesteps) of IMU measurements (\unit[180]{Hz}) $\mathbf m_t$ and visual embeddings (\unit[30]{Hz}) $\mathbf e_t$ as input. 
In contrast to approaches that jointly train vision and action module~\cite{laskin2020curl}, the vision encoder is frozen after the contrastive pretraining. 
We use two separate Temporal Convolutional Networks (TCN)~\cite{lea2016temporal} to encode the series of vision embeddings and IMU measurements. Then the output features from two TCNs are concatenated and fed into a two-layer MLP, which outputs the actions. Note that using two TCNs allows different update rates for IMU measurements and images. We define an action loss $\mathcal{L}_A$ (action error) that is the mean square error between the outputs of the teacher policy and the student policy
\begin{equation}
    \mathcal{L}_A(\bm{\theta}) = \lVert\bm{\pi}_{\textrm{student}}(\bm{\theta};~\bm{e}_t, \bm{m}_t) - \bm{\pi}_{\textrm{teacher}}(\bm{o}_t)\rVert
\end{equation}
\section{Experiments}
To achieve successful scene transfer for mobile robot navigation, we require a vision encoder that enables the action net to learn a meaningful sensorimotor policy that maps the environment observations to control commands. First, we briefly give implementation details and describe the training setup. Then we evaluate the performance of the contrastive encoder. Subsequently, we evaluate the performance of mobile robot control policies in a set of simulated and real-world experiments at the drone racing task. 

\subsection{Implementation Details and Model Training}
\subsubsection{Image data for vision encoder training} The images for vision encoder training are collected by rolling out the state-based teacher in four simulation environments (see Fig.~\ref{fig:method}a). These environments consist of a simple warehouse scene, a digital twin of the testing environment, an empty scene, and one realistic digital twin rendered using Blender~\cite{blender}. 
To replicate the sensor configuration on our quadrotor platform, the images are rendered using a grayscale fisheye camera model (double sphere model~\cite{usenko2018double}). In total, we collected 12000 images from three rollouts in four environments.

\subsubsection{Vision encoder training} We perform contrastive encoder training on the collected images as described in section \ref{subsec:contrastive}. The vision encoder consists of a backbone based on ResNet50~\cite{he2016deep} and a reprojection layer~\cite{chen2020simple} to generate 128-dimensional embedding from images of size 224$\times$224. For training, there are 24 negative samples and 12 positive samples associated with each image. For each anchor image, the positive samples are uniformly sampled within $[0, 0.05] \cup [0.95, 0,1]$ relative track progress, while negative samples are uniformly sampled within $[0.4, 0.6] $ relative track progress. The normalized track progress is a metric to describe the lap completion and is in the range of $[0, 1]$.
For the model training, we employ the Adam optimizer~\cite{adam} with a linearly decayed learning rate, initially set at 1e-4.

\subsubsection{Student policy learning} The student policy is trained via DAgger~\cite{ross2011reduction}. For imitation learning, we use a batch size of 512, and the training converges typically after collecting 500K data samples after around 100 epochs (about \unit[6]{h} on a desktop PC). We apply a linear decay on the learning rate, starting from 1e-3 and decreasing to 1e-5 at 50 epochs and remaining unchanged for the remainder of the training.

\subsection{Vision Embedding}
\subsubsection{Intra-scene consistency} For the action net to successfully learn a sensorimotor policy, we require descriptive, task-related, and informative vision embeddings. For the task of vision-based robots racing through a given track, we hypothesize that the intra-scene consistency is a suitable proxy measure: if two images are sampled closely to each other along the track (with a similar track progress), we would expect a very high similarity in the embedding space. Similarly, the further apart (in terms of progress along the racetrack), the lower the similarity should be. Intuitively, this facilitates action learning as close-by states require similar actions, whereas far-away states require different actions. For evaluation, we use 1000 images sampled around the teacher policy rollout trajectories.

\begin{figure}[t!]
    \centering
\tikzstyle{every node}=[font=\footnotesize]
\includegraphics[width=0.98\linewidth]{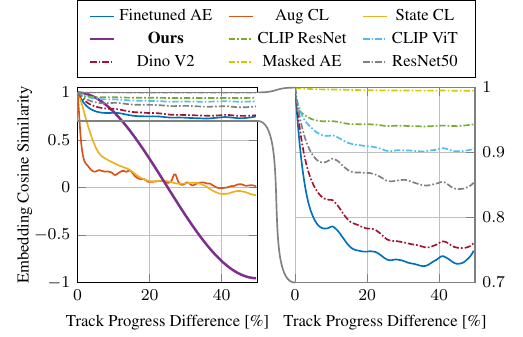}
\vspace*{-11pt}
\caption{This figure shows the intra-consistency of the different methods. Solid lines represent methods that have been finetuned and dash-dotted lines represent pretrained methods. One can clearly see that pretrained methods show a much more similar embedding, even at opposite ends of the track (50\%). Our proposed method produces an embedding that is very distinctive, as far points on the track have very dissimilar embedding.}
\label{fig:intraconsistency}
\vspace*{4pt}
\end{figure}

Fig.~\ref{fig:intraconsistency} shows the embedding cosine similarity as a function of track progress difference, where 0\% progress difference means the same location along the trajectory and 50\% progress difference is the opposite side of the track. The reported similarities are averaged values around the whole track. We can see that pretrained `world-model' like methods (dash-dotted lines) show very high similarities independent of track progress. This makes sense as most models are trained on very large datasets containing millions of different scenes\textemdash being on the opposite side of a racetrack in the same environment, thus only causing a small change in the embedding space. 

Methods finetuned on task-specific images from the racing environment (solid lines in Fig.~\ref{fig:intraconsistency}) generally show much less similarity for far away samples while retaining high similarity for close samples. Due to using 12 positive samples and augmenting the images at training time, our method achieves a much larger difference in embedding similarity than any other method.

\subsubsection{Inter-scene consistency} 
\begin{figure}[t!]
    \centering
    \includegraphics[width=\linewidth]{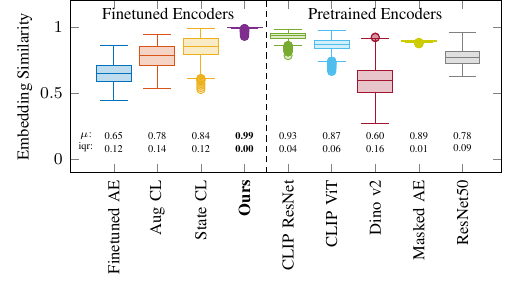}
\vspace*{-35pt}
\caption{This figure shows the inter-consistency of the different methods. This is a measure of how close the embeddings from different environments are together (higher is better). Our method clearly outperforms the baselines, as a high inter-consistency is desired while ensuring a low intra-consistency for far apart samples. The numbers below each bar indicate the mean embedding similarity $\mu$ and inter-quartile-range iqr.}
\label{fig:interconsistency}
\vspace*{-20pt}
\end{figure}

In the previous section we looked at intra-scene consistency as a proxy measure for the descriptiveness of the vision embedding. To enable scene transfer, however, the embedding does not only have to be descriptive but also invariant to the scene in which the task is performed. In Fig.~\ref{fig:interconsistency}, we analyze how similar embeddings are if the robot's pose remains unchanged but the environment changes. Again, 1000 poses around the track are sampled, and the cosine similarity of the embeddings is compared for each pose.

First, one notices that there is no clear trend between pretrained and finetuned methods. Our method clearly outperforms all other methods with both the lowest interquartile range (iqr) and the highest average similarity of 0.99. Other methods like the masked autoencoder (Masked AE) or CLIP also perform very well with high similarity and low variance.
To understand the true performance, however, we have to combine the results of intra-scene and inter-scene consistency: while the masked AE shows great results in the inter-scene consistency, it shows no distinctiveness in the intra-scene consistency, meaning that it just always generates similar embeddings. The contrastive learning baselines (Aug CL and State CL) performed well in terms of intra-scene distinctiveness but showed very high variance in the inter-scene consistency. Only our method is able to provide a task-related descriptive embedding while being scene-invariant. 

\subsection{Embedding visualization}
\begin{figure}[t!]
    \centering
\includegraphics[width=\linewidth]{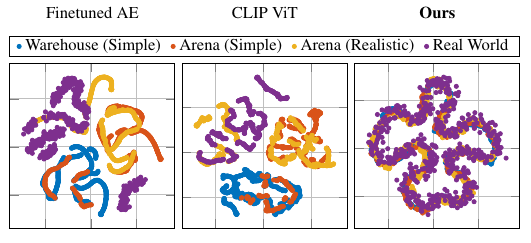}
\hspace*{-2.5mm}
\vspace*{-10pt}
\caption{t-SNE Visualizations of the embeddings for different environments. The illustration graphically summarizes what can be read quantitatively from Fig.~\ref{fig:intraconsistency} and Fig.~\ref{fig:interconsistency}: only our approach produces distinctive, task-related embeddings that are similar across all environments.}
\label{fig:tnse}
\vspace*{-14pt}
\end{figure}

We apply a t-SNE dimensionality reduction~\cite{van2008visualizing} to visualize the embeddings from four distinct environments in a 2D space, as depicted in Figure~\ref{fig:tnse}. This visualization shows graphically what the previous section described: Firstly, our approach effectively encodes images within each environment into coherent and continuous representations. 
Secondly, for our approach, the vision embeddings across multiple environments are extremely similar (on top of each other). For both types of baselines, we observe somewhat coherent clustering (lines of the same color are mostly continuous, but the environments are vastly different from each other.

\subsection{Action Learning}
\label{sec:act_learn}
Our proposed vision encoder has achieved very good results in terms of consistency. We now turn towards the action learning to analyze which embedding is suitable for agile, vision-based flight. In this section, we use a fixed vision encoder that has been trained on all simulation environments (warehouse, no background, digital twin), but has never seen real-world data.

\begin{table}[t]
    \centering
    \caption{\vspace*{6pt}\parbox{1\linewidth}{
    \textnormal{Action Error of the different methods evaluated in four different settings: the training environment, a simple unseen test environment, a highly realistic (unseen) blender render using a digital twin of the real world, and real-world flight data. Our method outperforms all baselines.}}}
    \vspace*{-7pt}
    \setlength{\tabcolsep}{2pt}
    \label{tab:openloop}
    \begin{tabularx}{1\linewidth}{l|x{1.0cm}|CCC|C}
    \toprule
    \textbf{Methods}  & \centering \textbf{Custom} \newline \textbf{Dataset}& \textbf{AE Train (Simple)}& \textbf{AE Test (Simple)} & \textbf{AE Test (Realistic)} & \textbf{AE Real World }\\
    \midrule
    CLIP ResNet        & \xmark & 0.016 & 0.043 &  0.142 & 0.116\\ 
    CLIP ViT~\cite{radford2021learning}   & \xmark& 0.019 &0.057  &  0.154 & 0.115\\
    MAE~\cite{mae2022kh} & \xmark& 0.019  & 0.063 &  0.217 & 0.136\\
    Resnet~\cite{he2016deep}           & \xmark&  0.018&0.071&  0.112 & 0.087\\
    Dino V2~\cite{oquab2023dinov2}               & \xmark & 0.019  &0.062  &  0.124 & 0.098\\
    Autoencoder       & \cmark & 0.015  & 0.026& 0.068 & 0.088\\
    Aug CL \cite{laskin2020curl}      & \cmark & 0.011 & 0.036  & 0.091 & 0.078\\
    State CL \cite{eysenbach2022contrastive}  & \cmark & 0.009 & 0.031 & 0.069 & 0.073 \\
    \midrule
    Ours w/o pose  &\cmark & 0.010 & 0.022 & 0.041 & 0.043\\
    \textbf{Ours}  &\cmark  & \textbf{0.008}  &  \textbf{0.012}& \textbf{0.039} &\textbf{0.036}\\
    \bottomrule
\end{tabularx}
\vspace*{-12pt}
\end{table}

In the first set of simulation and real-world experiments, we record a rollout of the teacher policy and then evaluate the action error of the student policy w.r.t. the teacher policy. We use the teacher policy to generate the rollout to ensure fairness and evaluate all approaches on the same data. The student policies are trained via DAgger~\cite{ross2011reduction} in simulation and have only seen the warehouse and no-background environment. The digital twin, as well as real-world data, is not used during action-net training. For drone dynamics, we use a very accurate dynamics model~\cite{bauersfeld2021neurobem, foehn2022agilicious}. The evaluation is performed in four different settings: in the training environment, in the unseen digital-twin environment (both with the low-fidelity renderer and the realistic blender renderer), and on data collected in the real world. The results are summarized in Table~\ref{tab:openloop}. In all test cases, our method achieves the lowest action error, with the difference becoming more pronounced when scene transfer matters: on the unseen digital twin, we outperform the baselines by a factor of two. In the real world, all methods degrade significantly, but our method still outperforms the next-best by 25\%. 

\begin{table}[b!]\setlength{\hfuzz}{1.1\columnwidth}
\vspace*{-16pt}
\begin{minipage}{\textwidth}
    \centering
    \caption{\vspace*{6pt}\parbox{1\linewidth}{
    \textnormal{Success rate (SR), Action Error, and Average Gate Passing (AGP) of the different methods when evaluated in closed-loop operation for the drone racing task. When the control policy is deployed in an environment it has been trained in, all methods perform very well. When deployed in an environment that is unseen during the training of the action net, only our approach is able to perform the task, highlighting its ability to transfer.}}}
    \vspace{-7pt}
    \label{tab:closedloop}
    \begin{tabularx}{1.0\linewidth}{x{2cm}|x{1.0cm}|cCc|cCc|cCc}
    \toprule
    \multirow{2}{*}{\textbf{Methods}}  & \textbf{Custom} & \multicolumn{3}{c|}{\textbf{Train (Simple)}}&\multicolumn{3}{c|}{\textbf{Test (Simple)}}&\multicolumn{3}{c}{\textbf{Test (Realistic)}} \\
    & \bf Dataset & SR [\%]$\,\uparrow$ & Act Err$\,\downarrow$ & AGP\,$\uparrow$ & SR [\%]\,$\uparrow$   &  Act Err\,$\downarrow$   & AGP\,$\uparrow$  & SR [\%]\ $\uparrow$  &  Act Err\,$\downarrow$  & AGP\,$\uparrow$  \\
    \midrule
    CLIP Rn50~\cite{radford2021learning}        & \xmark &93.8 & 0.014& 9.42$\pm$0.20& 0.0 & 0.078 &2.85$\pm$0.48& 0.0 & 0.194&1.55$\pm$0.16 \\ 
    CLIP ViT~\cite{radford2021learning}  & \xmark& 95.3 & 0.015 &9.29$\pm$0.43& 0.0 & 0.052&2.93$\pm$0.29  & 0.0& 0.203&1.46$\pm$0.16\\
    MAE~\cite{mae2022kh} & \xmark& 89.1 & 0.017 &9.39$\pm$0.28&0.0 & 0.043&3.02$\pm$0.27& 0.0 & 0.210&1.32$\pm$0.14 \\
    Resnet~\cite{he2016deep}          & \xmark&95.3 & 0.013&9.51$\pm$0.16& 0.0 & 0.072&2.83$\pm$0.46& 0.0 & 0.112&2.08$\pm$0.33  \\
    Dino V2~\cite{oquab2023dinov2}               & \xmark & 95.3 & 0.015&9.05$\pm$0.19 &0.0 & 0.058&2.98$\pm$0.30& 0.0 & 0.153 &1.54$\pm$0.15 \\
    Autoencoder       & \cmark & \textbf{98.4} & 0.015&9.31$\pm$0.38& 0.0 & 0.136 &1.67$\pm$0.19& 0.0 & 0.139&1.67$\pm$0.19 \\
    Aug CL \cite{laskin2020curl}      & \cmark &96.8 & 0.011&9.51$\pm$0.44 & 0.0 & 0.091&3.11$\pm$0.31& 0.0& 0.124&1.91$\pm$0.21  \\
    State CL \cite{eysenbach2022contrastive}  & \cmark & \textbf{98.4} & \textbf{0.010}&\textbf{9.82}$\pm$0.32&0.0 & 0.040&3.68$\pm$0.32 & 0.0 & 0.082&2.45$\pm$0.12  \\
    \midrule
    Ours w/o pose & \checkmark & 96.8 & 0.013 & 9.57$\pm$0.42 & 48.4 & 0.035 &5.01$\pm$0.23 & 31.3 & 0.073 & 4.43$\pm$0.10 \\
    \textbf{Ours}  &\cmark  & \textbf{98.4} & 0.012 &9.65$\pm$0.20  &\textbf{64.1} & \textbf{0.023}&\textbf{7.42}$\pm$0.42&\textbf{43.8} & \textbf{0.047}&\textbf{6.03}$\pm$0.56\\
    \bottomrule
\end{tabularx}
\vspace*{-10pt}
\end{minipage}
\end{table}

In a second set of experiments, we evaluate the closed-loop control performance of all methods in the same simulation environments as already presented above. We do not show real-world results as our current network is not running fast enough on the onboard computer Nvidia Jetson TX2 to stabilize the vehicle (requires at least 100Hz control frequency). The results are summarized in Table~\ref{tab:closedloop}. We define the evaluation metrics success rate (SR) as the ratio of successful tests over the total number of tests, where each of the 64 tests starts from a different position that is uniformly sampled from the predefined box, and the number of test steps covers two complete laps. Each test is counted as a ``success" only when the robot survives the whole testing episode without crashing. In the training environment, all methods perform adequately, and the differences in terms of action error (AErr) and SR are small. When the methods are deployed on the unseen digital twin of our testing facility, all methods except ours fail while we retain at least a 44\% SR in the highly realistic simulator.
Meanwhile, we also demonstrate that our approach could generalize to unseen random scene/brightness/blurriness changes during closed-loop deployment. We refer the reader
to the accompanying video to visualize these
experiments.
\subsection{Beyond Drone Racing}
To demonstrate that our approach can be applied beyond vision-based agile flight tasks, we ablate the importance of pose annotations for the contrastive encoder training. 
Instead of privileged pose information, we use only time-series information during the encoder training. This kind of information is available for all sequential tasks and equally applies to drone racing as it does to open a door. Additionally, instead of the adaptive temperature shown in Eq.~(\ref{eq:similarity}), a constant one of 0.5 is used, as our adaptive temperature formulation is based on the poses of the individual samples as shown in Eq.~(\ref{eq:similarity}). 
The results shown in Table~\ref{tab:openloop} (`Ours w/o pose') demonstrate that the privileged information helps the overall performance of our approach. Nevertheless, relying only on time-series information to train the contrastive encoder, we still outperform all baselines and are (on real-world data) only 10\% worse in terms of action error compared to a privileged encoder training. This strongly suggests our contrastive training pipeline has implications for general mobile robotics tasks beyond drone racing.
\enlargethispage{-15.65\baselineskip}
\section{Conclusion}
We presented a contrastive learning-based approach to facilitate scene transfer for vision-based agile flight. By using a novel adaptive loss formulation, the network was able to learn a task-specific embedding that is highly descriptive and consistent across many scenes. These properties set it apart from large `world-model type', pre-trained encoders like CLIP~\cite{radford2021learning} or Dino v2~\cite{oquab2023dinov2}. We evaluated our method on seen and unseen environments and found that it outperforms all baselines in terms of action error and closed-loop experiment success rates. Lastly, we showed evidence that our contrastive training pipeline can not only be used for drone racing but generalized to vision-based sequential tasks. Future work could include more prior knowledge of the physical world to further enhance policies' robustness and safety~\cite{romero2024icra}.

\clearpage
{\small
\balance
\bibliographystyle{IEEEtran}
\bibliography{all}
}

\end{document}